\pdfoutput=1

\documentclass[11pt]{article}

\usepackage[final]{acl}

\usepackage{times}
\usepackage{latexsym}

\usepackage[T1]{fontenc}

\usepackage[utf8]{inputenc}

\usepackage{microtype}

\usepackage{inconsolata}

\usepackage{graphicx}

\usepackage{comment}
\usepackage[utf8]{inputenc} %
\usepackage[T1]{fontenc}    %
\usepackage{hyperref}       %
\usepackage{url}            %
\usepackage{booktabs}       %
\usepackage{amsfonts}       %
\usepackage{nicefrac}       %
\usepackage{microtype}      %
\usepackage{xcolor}         %
\usepackage{arydshln}
\usepackage{pifont}
\usepackage{multirow}
\usepackage{xspace}
\usepackage{graphicx}
\usepackage{cleveref}
\usepackage{pgfplots}
\pgfplotsset{compat=1.18}

\usepackage{colortbl}
\colorlet{tableheadcolor}{gray!75}
\colorlet{tablerowcolor}{gray!10}
\newcommand{\rowcol}{\rowcolor{tablerowcolor}} %
\usepackage{siunitx}

\usepackage{algorithm}
\usepackage{algpseudocode}

\newcommand{\ssvp}{SSVP-SLT\xspace}

 \title{SignMusketeers: An Efficient Multi-Stream Approach\\for Sign Language Translation at Scale}

\author{Shester Gueuwou, Xiaodan Du, Greg Shakhnarovich, Karen Livescu \\ 
\small TTI-Chicago \\  
\small{\{shesterg,xdu,greg,klivescu\}@ttic.edu}\\
\url{http://signmusketeers.pals.ttic.edu}
}

\begin{document}
\maketitle
\begin{abstract}
A persistent challenge in sign language video processing, including the task of sign to written language translation, is how we learn representations of sign language in an effective and efficient way that preserves the important attributes of these languages, while remaining invariant to irrelevant visual differences. Informed by the nature and linguistics of signed languages, our proposed method focuses on just the most relevant parts in a signing video: the face, hands and body pose of the signer. However, instead of fully relying on pose estimation from off-the-shelf pose tracking models, which have inconsistent performance for hands and faces, we propose to learn a representation of the complex handshapes and facial expressions of sign languages in a self-supervised fashion. Our approach is based on learning from individual frames (rather than video sequences) and is therefore much more efficient than prior work on sign language pre-training. Compared to a recent model that established a new state of the art in sign language translation on the How2Sign dataset, our approach yields similar translation performance, using less than 3\% of the compute. 
\end{abstract}

\section{Introduction}
\label{intro}

\begin{figure*}[t!]
    \centering
    \includegraphics[width=\textwidth]{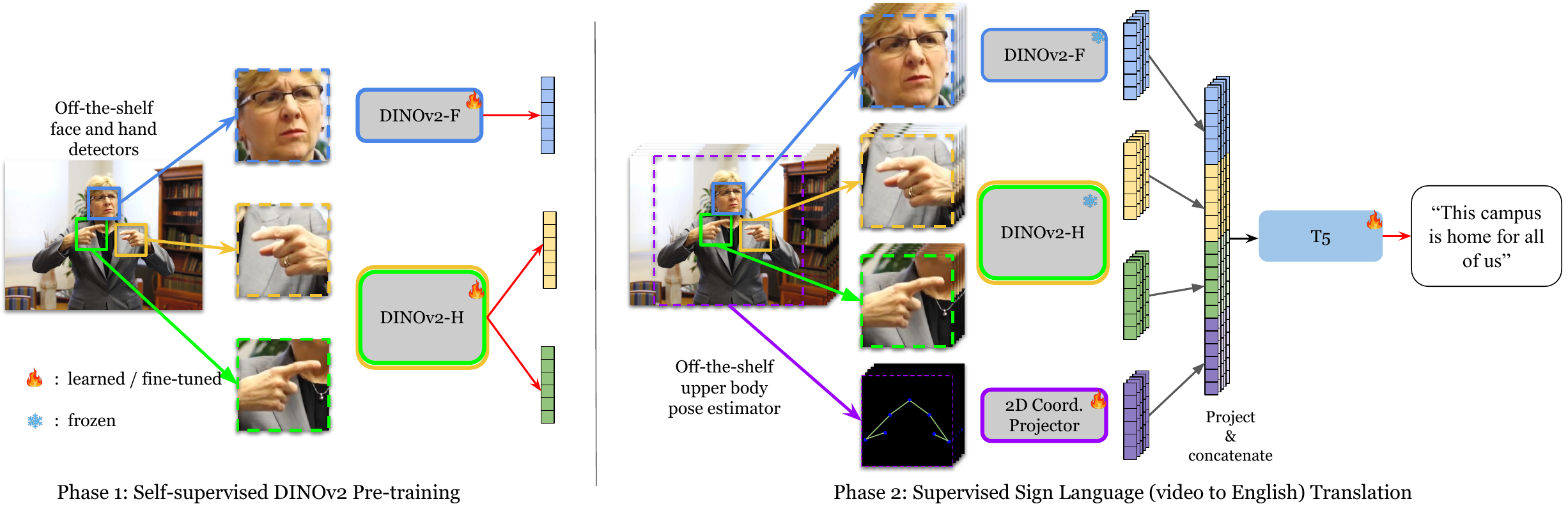}
    \caption{Overview of our approach to sign language translation. We parse every frame of the signing video with off-the-shelf face and hand detectors. (a) In phase 1 (left) we start from pre-trained DINOv2 visual feature extractors and continue training them with a DINO loss on cropped face boxes and hand boxes, producing two separate DINOv2s (DINOv2-F for the face and DINOv2-H for the hands). This stage is purely self-supervised from random video frames; see also Fig.~\ref{fig:pipeline} for more detail. (b) In phase 2 (right), fixing the two pre-trained feature extractors, we add a (learned) feature extractor for coarse body pose estimated by an off-the-shelf method \cite{lugaresi2019mediapipe}, concatenate and project the features for each frame, and fine-tune a T5 model mapping the resulting sequence of frame features to English text. This stage is supervised by video clips paired with translations.}
    \label{fig:overview}
\end{figure*}

Recent work on sign language processing spans human-computer interaction \cite{zafrulla2010american, bragg2021asl}, computer vision~\cite{varol2021read,sandoval2023self}, and  natural language processing (NLP) \cite{yin-etal-2021-including,muller-etal-2022-findings} research. The nature of signed languages, which involve the use of manual features (handshape, orientation, location, and movement) and non-manual features (facial expressions, head movements, body pose), presents challenges for machine learning models \cite{bragg2019sign}.  In particular, how to effectively and efficiently represent signed languages while preserving their inherent attributes remains a persistent challenge.

Our focus is on the task of translation from sign language video to a written language (sign language translation, or SLT). This is one of the most practically important tasks, necessary to bridge (part of) the communication gap between Deaf and hard of hearing (DHH) populations and hearing populations \cite{fox2023best}. Recent work~\cite{rust2024towards} proposed a self-supervised video pre-training approach to handle sign language translation, which achieved state-of-the-art performance on the How2Sign dataset of American Sign Language (ASL) to English translation~\cite{duarte2021how2sign}. The intuition behind this approach is to extend the pre-training of a strong video model (in this case, Hiera \cite{ryali2023hiera}) with a large-scale unannotated set of sign language videos (in this case from YouTube-ASL~\cite{uthus2023youtube}) and use this extended pre-trained model as a feature extractor for the supervised translation task. However, this pre-training is extremely costly: Its longest pre-training run uses 64 A100 80GB GPUs for 14 days, making the approach infeasible for many researchers and practitioners.

The approach in~\citet{rust2024towards} implicitly treats sign language sequences like any other (long) videos. But signed languages are, first and foremost, languages, and like any other language, they possess linguistic properties that  may provide an inductive bias about the more important aspects of the video \cite{brentari1998prosodic,Sutton-Spence_Woll_1999}. In this work, we ask the question: \textit{Can we infuse basic linguistic properties of signed languages into self-supervised pre-training to develop a scalable compute-friendly approach?}

In the context of signed language processing, the use of off-the-shelf human pose estimator models \cite{cao2017realtime,lugaresi2019mediapipe} has been one of the most common ways of incorporating the linguistic constraints of sign language into models. The intuition of using pose estimation is that it removes irrelevant features that do not affect the meaning of signs, such as body shape and visual background, and therefore focuses entirely on the linguistically relevant aspects of hand, face and body pose \cite{de2023towards}. However, this pose-based approach has several limitations that make it sub-optimal for capturing the details of signed languages, particularly in the representation of hands \cite{moryossef2021evaluating} and faces \cite{kuznetsova2024testing}. First, the human pose estimator models used in existing methods~\cite{cao2017realtime,lugaresi2019mediapipe,mmpose2020} are typically trained on everyday handshapes, which are often less complex than the handshapes found in signed languages. Second, human pose estimators are unreliable in capturing crucial non-manual components which are essential for signed languages, such as eye gaze.\footnote{For instance, in British Sign Language, the main difference between the signs for "God" and "Boss" lies in the eye gaze, which existing human pose estimators do not capture~\cite{Sutton-Spence_Woll_1999}.}

Our approach is inspired by the multi-stream/multi-channel property of signed languages (\S \ref{sec:multi_channel_sign_language_processing}); that is, the fact that they consist of a combination of actions performed largely independently by multiple body parts (channels). Specifically, our proposed method (\S \ref{sec:method}, see overview in Fig.~\ref{fig:overview}) focuses on the most relevant parts of a signing video---the face, hands, and body pose of the signer (\S \ref{sec:data_preprocessing})---to handle sign language translation at scale (\S \ref{sec:sign_language_translation_at_scale}). Instead of fully relying on off-the-shelf human pose estimators, we propose to learn representations of handshapes and facial expressions directly from signing videos using self-supervised learning (\S \ref{sec:self-supervised_sign_components_pre-training}, \ref{sec:face_hands_and_body_posture_representation_learning}.) Our method captures the intricacies of handshapes and facial expressions for the supervised training stage (\S \ref{sec:supervised_sign_language_translation}), without the need for extensive pre-training data or computational resources. This allows us to overcome the limitations of human pose estimators and preserve the crucial linguistic information conveyed through handshapes, eye gaze, and facial expressions in signed languages. We name our approach SignMusketeers: Like the heroes of Dumas' books \cite{dumas1894works}, three image channels (face and two hand boxes) join forces with a fourth companion (pose features) in the quest for glory (accurate sign language translation).

We conduct experiments (\S \ref{sec:experiments}) on How2Sign and find that our approach achieves competitive performance while using a smaller model (in terms of number of parameters), with 41x less pre-training data and 160x fewer pre-training iterations (\S \ref{sec:results}) , using roughly 3\% of the compute resources of the previous state-of-the-art approach~\cite{rust2024towards}. We also provide ablation experiments showing the value of individual design decisions (\S \ref{sec:ablation}).

\section{Related Work}
\label{sec:related_works}

\subsection{Sign Language Translation at Scale}
\label{sec:sign_language_translation_at_scale}
Until very recently, the lack of large-scale datasets has been a major obstacle in advancing sign language translation. Most research has been conducted on small datasets such as PHOENIX14T \cite{camgoz2018neural}, with only 9 hours of content and a modest vocabulary size of 3,000. While fairly high BLEU\footnote{Unless specified otherwise, BLEU means BLEU-4 scores, computed with \texttt{sacrebleu} \cite{post-2018-call} \texttt{version : BLEU+c.mixed+\#.1+s.exp+tok.13a+v.1.4.1.}} scores (>20) have been reported on this dataset, translation of more realistic video is far more challenging.
Recent efforts to create larger datasets, such as BOBSL ($\sim$1,500 hours) \cite{Albanie2021bobsl}, OpenASL ($\sim$300 hours) ~\cite{shi-etal-2022-open}, JWSign ($\sim$2,500 hours) \cite{gueuwou-etal-2023-jwsign}, and the SRF corpus ($\sim$400 hours) \cite{muller-etal-2023-findings}, have revealed the difficulty of the task, with BLEU scores only around 2-7.

In this study, we do not include common benchmark datasets like PHOENIX14T \cite{camgoz2018neural} and CSL-Daily \cite{zhou2021improving} due to two main reasons: (i) The goal of our study is to investigate sign language translation at \emph{scale}. We focus on American Sign Language (ASL) because it has easily accessible large-scale datasets for pre-training and smaller datasets for fine-tuning. At the time of this study, there were no easily accessible large-scale datasets for German Sign Language or Chinese Sign Language that would support our approach. (ii) Using YouTube-ASL \cite{uthus2023youtube} for pre-training and How2Sign \cite{duarte2021how2sign} for fine-tuning follows established precedent in prior work (mentioned below), enabling direct comparisons with our method.

\citet{uthus2023youtube} used the YouTube-ASL dataset for large-scale training of ASL translation models. By fine-tuning a T5 \cite{raffel2020exploring} model on YouTube-ASL, then fine-tuning it on the smaller dataset How2Sign \cite{duarte2021how2sign}, the authors achieved a BLEU score of 12.39. The input to the T5 model consisted of selected human poses obtained from the off-the-shelf human pose estimator MediaPipe \cite{lugaresi2019mediapipe}.

Building upon this paradigm, \citet{rust2024towards} further improved performance on How2Sign by training a video encoder on YouTube-ASL initialized from a self-supervised video model (Hiera-Base) pre-trained as a masked autoencoder. However, this approach is computationally prohibitive, requiring 64 A100 80GB GPUs for 14 days for a single training run. 
The authors found that the good results indeed depend on these large compute requirements: Significantly reducing the number of video frames ingested by the encoder from 128, or the number of pre-training iterations from 800 epochs, greatly reduced performance. One key feature of this method is its privacy-awareness through face blurring.\footnote{The details of the blurring approach in~\cite{rust2024towards}, that uses an internal software tool, have not been published, making it irreproducible.}
However, the face is an important non-manual cue that helps disambiguate some statements.
We also note that face blurring may not be sufficient for preserving privacy, especially with large-scale datasets~\cite{oh2016faceless}. 

We propose an alternative approach that focuses on embedding fine-grained handshapes and facial expressions using an image encoder with a smaller ViT backbone. Our encoder takes just a single frame at a time and requires much lower training time and compute resources.

\subsection{Multi-Channel Sign Language Processing}
\label{sec:multi_channel_sign_language_processing}

Signed languages are inherently multi-channel, employing a combination of manual features (handshapes, orientation, location, and movement) and non-manual features (facial expressions, head movements, and (upper) body pose) to convey meaning \cite{sandler2006sign,Pfau2010NonmanualsTG, brentari2019sign}. Multi-channel processing aims to capture and integrate these diverse sources of information for various tasks, such as sign language recognition, translation, and generation. The concept of tackling (American) sign language processing through a multi-channel approach was first introduced in the early 2000s \cite{vogler2001framework}, inspired by linguistic evidence that American Sign Language can be modeled, at least partially, as a combination of independent channels \cite{liddell1989american}. Over the years, several other approaches have used multi-channel ideas for sign language recognition \cite{holden2005australian,pu2016sign} and later translation \cite{camgoz2020multi,zhou2021spatial,shi-etal-2022-open} and production \cite{saunders2020adversarial, tornay2020phonology} tasks, although the specific channels and how they are used varies. 
One common characteristic in these approaches is that the feature extractors for the different components were not learned specifically for sign languages. This is an example of a general issue in sign language research that the methods are not sufficiently adapted to the needs of these languages~\cite{fox2023best,desai2024systemic}. %

\subsection{Face, Hands, and Body Pose Representation Learning}
\label{sec:face_hands_and_body_posture_representation_learning} 
To achieve our goal of learning semantically meaningful multi-channel features independently of irrelevant visual details, we draw inspiration from prior work related to facial expression representation learning, hand pose (or shape) estimation, and body pose learning.   

\paragraph{Facial expression.}  One approach for face analysis is~\citet{da2020recognition}, which investigates the recognition of affective and grammatical facial expressions in Brazilian Sign Language (Libras). The authors utilize a combination of geometric features, such as facial landmarks, and appearance features to represent facial expressions. Another approach is MARLIN \cite{cai2023marlin}, a masked autoencoder for facial video representation learning, which is effective on various facial analysis tasks, including facial expression recognition. \citet{gao2024self} propose a self-supervised learning approach for facial representation learning with facial region awareness. This method leverages the structure of the human face by dividing it into regions, such as eyes, nose, and mouth. The authors utilize BYOL \cite{grill2020bootstrap}, a popular self-supervised learning framework based on instance discrimination. Our approach shares some ideas with the work of \citet{gao2024self}; however, instead of using BYOL, we employ DINOv2 \cite{oquab2024dinov}, a state-of-the-art visual self-supervised learning framework that builds upon a similar principle of instance discrimination as BYOL.

\paragraph{Hand shape and orientation.} 
DeepHand \cite{koller2016deep} is a convolutional neural network (CNN) approach for hand shape classification in continuous sign language video streams. It addresses the challenge of weakly labeled data by proposing a training strategy that exploits the temporal coherence of hand shapes within a sign. 
\citet{zimmermann2021contrastive} propose a contrastive representation learning approach for hand shape estimation, which learns hand shape representations by contrasting positive and negative pairs of hand images in a self-supervised manner, using a novel loss function that encourages invariance to changes in viewpoint, articulation, and lighting conditions. FineHand \cite{santhalingam2020finehand} is a deep learning approach specifically designed for American Sign Language (ASL) recognition. Our approach draws inspiration from these studies and aims to learn hand shape representations in a self-supervised manner using DINOv2, which enables us to capture the fine-grained details of hand shapes and orientation without relying on explicit annotations. We note also that \citet{wong2024signgpt} used DINOv2 as a feature extractor backbone for sign language translation.

\paragraph{Upper body pose.}  Compared to analysis of the fine-grained gestures of the face and hands, techniques for general (global) human pose estimation~\cite{hachiuma2023unified,yan2023skeletonmae} are more mature and robust.  In our approach, we therefore simply utilize an off-the-shelf human pose estimation model, MediaPipe~\cite{lugaresi2019mediapipe}, to represent the body pose of a signer.

\section{Method}
\label{sec:method}
Below we first describe the data preprocessing (frame parsing) procedure, then the self-supervised pre-training of feature extractors, and finally the supervised learning of the ASL video to English translation system.

\subsection{Frame Parsing}
\label{sec:data_preprocessing}

Our frame parsing pipeline extracts and normalizes the relevant regions of interest (ROIs) from the sign language video frames, focusing on the face, left hand, right hand, and upper body pose. Each of these four components is mapped to a feature vector (channel); concatenating and projecting the four channels for each frame yields the frame vector, which is then fed to a sequence model for translation. We use the MediaPipe Holistic framework \cite{lugaresi2019mediapipe} to extract face, hand, and pose landmarks from the video frames.

\vspace{.1in}
\noindent\textbf{Face cropping} To extract the face ROI, we first determine the smallest square bounding box that can fit all the face landmarks while preserving the aspect ratio of the face. This initial bounding box is then scaled up by a factor of 1.2 in each dimension to compensate for any parts of the face that might have been missed. 

In cases where face landmarks are not detected, we estimate the face region using the upper body pose landmarks (indices 0 to 10). The bounding box is adjusted to ensure it fits within the frame boundaries.

\vspace{.1in}
\noindent\textbf{Hand cropping} For the hand ROIs, we follow a similar approach to the face ROI extraction when hand landmarks are available: We determine the smallest square bounding box that can fit all the hand landmarks while preserving the aspect ratio of the hand and scale it by a factor of 1.2. 

In cases where hand landmarks are not detected, we estimate the hand regions using the few finger pose landmarks (indices 17, 19, 21 for the left hand fingers and 18, 20, 22 for the right hand fingers. See \Cref{fig:mediapipe_indices} in \Cref{sec:mediapipe_pose_indices} for an ilustrative diagram). Although these are often inaccurate, they provide a good estimate of the hand location(s). We then create a square bounding box of the same size as the face bounding box with its center at the mean of the relevant pose landmarks. 

To handle the occasional cases where the MediaPipe hand landmark detector returns erroneous values when the hand is not in the frame, we adjust the bounding boxes to maintain temporal consistency across the channels. Specifically, we use two strategies: shifting the bounding box inward to keep the hand within the frame or using the last previously detected hand ROI before it went out of the frame.

The extracted face and hand ROIs are then resized to a fixed 224x224 pixels using bicubic interpolation while preserving the aspect ratio. This resizing step ensures consistent input dimensions for the self-supervised learning models in the next stage.

\vspace{.1in}
\noindent\textbf{Upper body pose normalization} We extract body poses from the relevant upper body landmarks, and normalize them to encourage invariance to position and scale differences.  Specifically, we extract MediaPipe human poses for the nose (index 0), left shoulder (index 11), right shoulder (index 12), left elbow (index 13), right elbow (index 14), left wrist (index 15), and right wrist (index 16). We assume that these seven landmarks are enough to capture the essential components of the upper body pose needed to recognize movements and spatial positions of the hands and face with respect to each other, leaving the finer-grained hand pose and facial expression to the other channels (face and hand ROIs).

Next, we define a normalized signing space based on the signer's body proportions, similarly to~\citet{bohavcek2022sign}. We define the head unit as the distance between the left and right shoulders divided by 2. The signing space width is set to 6 times the head unit, and the signing space height is set to 7 times the head unit. The signing space bounding box is determined using the left eye and nose landmarks as reference points.

Finally, we normalize the seven pose coordinates by scaling the bounding box of the signing space to unit width/height with center at (0.5,0.5). The normalized coordinates are then flattened into a (14-dimensional) vector.

To handle cases where the pose landmarks are not detected in a frame, we employ a strategy similar to the one used for the hands: If the pose landmarks are not detected and there is a pose available from a previous frame, we use the previous pose for the current frame. If there is no previous pose available, we create a placeholder array of negative values to indicate missing data \cite{uthus2023youtube}.

\subsection{Self-Supervised Sign Components Pre-training}
\label{sec:self-supervised_sign_components_pre-training}

\begin{figure*}[t!]
    \centering
    \includegraphics[width=.95\textwidth]{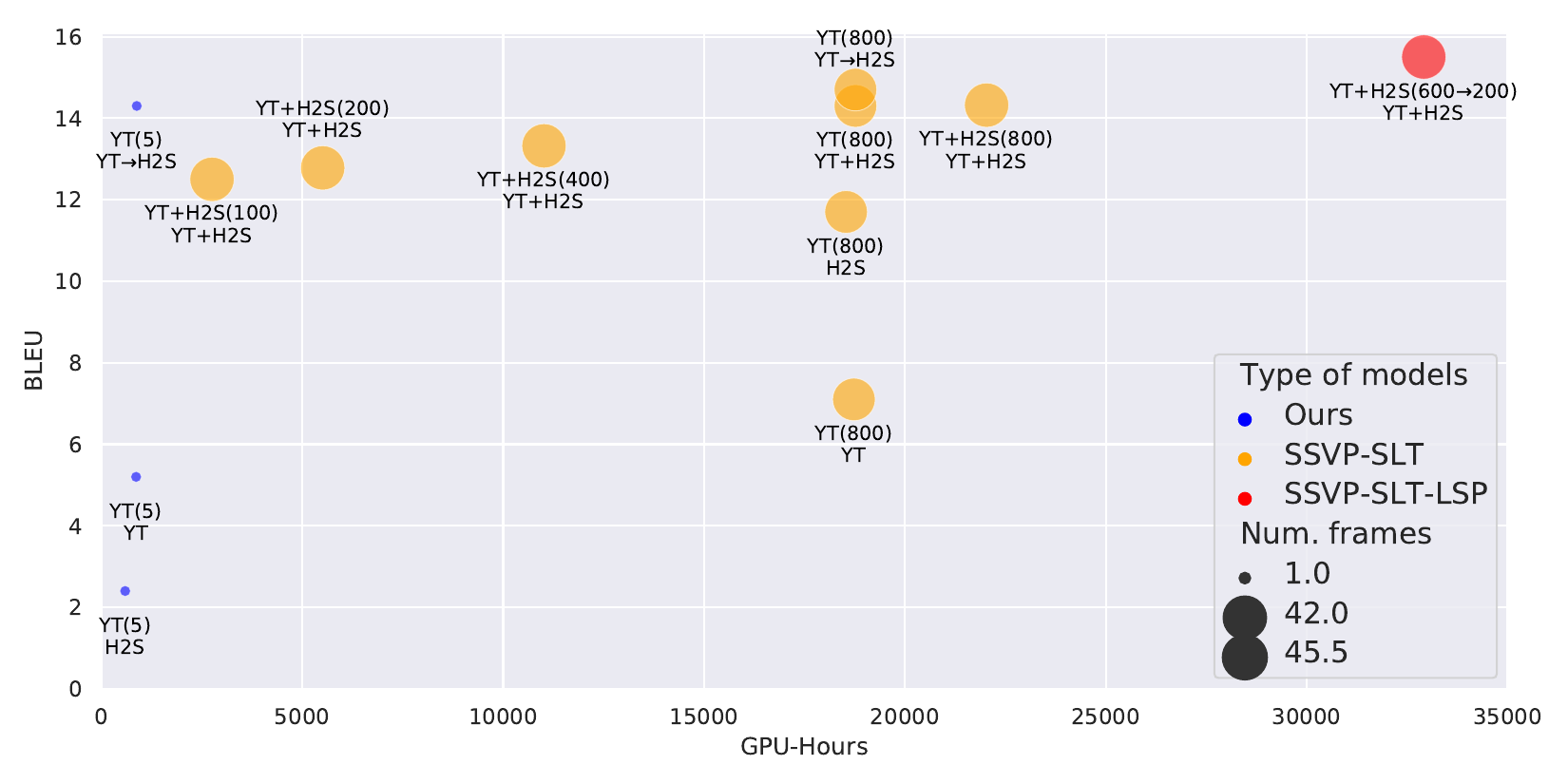}
    \caption{Comparison of data and computation usage between SignMusketeers (Ours) and \citet{rust2024towards}. Horizontal axis: GPU-Hours for the entire training schedule, including self-supervised training and supervised training. Vertical axis: BLEU score. Bubble size: number of frames (in millions) used during the pre-training stage. Labels: the first line is the pre-training protocol and the second line is the supervised training protocol. The number in parentheses is the number of pre-training epochs. YT: YouTube-ASL, H2S: How2Sign; \textup{X}$\to$\textup{Y} means train on \textup{X} then fine-tune on \textup{Y}; \textup{X}+\textup{Y} means train on the union \textup{X}$\cup$\textup{Y}. Note: GPU-Hours for~\citet{rust2024towards} is computed based on Section C.3 of~\citet{rust2024towards}.}
    \label{fig:bubble}
\end{figure*}

\begin{figure}
    \centering
    \includegraphics[width=\columnwidth]{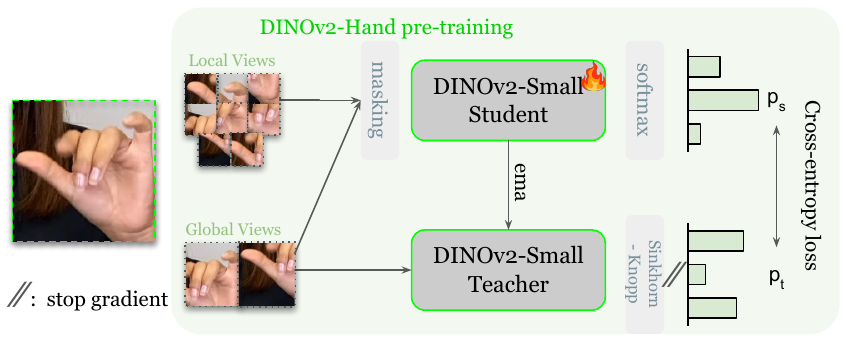}
    \caption{Self-supervised pre-training of DINOv2 on hand crops (stage 1 of our approach), yielding the hand-specific DINOv2 feature extractor. We pool the right and left hand boxes. We repeat this step separately for face boxes, yielding the face-specific feature extractor.}
    \label{fig:pipeline}
\end{figure}

Our method has two training phases. The first stage is self-supervised, and aims to produce encoders that specialize in sign language facial expressions and hand gestures. Using the DINOv2-Small architecture \cite{oquab2024dinov,darcet2024vision}, we pre-train the two encoders separately. We initialize ViT-small {\it student} and {\it teacher} backbones (due to computational constraints, we could not use a ViT-Base backbone as in \citet{rust2024towards}) with the teacher weights from the original dinov2\_vits14\_reg checkpoint, while the linear heads are randomly initialized. 
 
We largely follow the training protocols of the DINOv2 paper with the recommended 4 registers \cite{darcet2024vision}. The input face/hand images are randomly transformed into 2 global and 8 local views, using a scale of 0.5 to 1.0, and 0.25 to 0.5, respectively. Then, all views (both global and local) are passed to the student network. The student features are normalized with a Softmax to obtain the score vectors $p_s$. The teacher network, on the other hand, only accepts global views as input. We use the Sinkhorn-Knopp centering algorithm~\cite{caron2020unsupervised} on the features from the teacher network to obtain $p_t$. We compute cross-entropy loss between $p_s$ and $p_t$ and use it to update the student network. The gradient backpropagation is disabled for the teacher network. The weights of the teacher network are updated with an exponential moving average (ema) of the student network weights. In addition, DINOv2 also masks out random patches of the input views to the student and computes a patch-level cross-entropy loss (iBOT loss~\cite{zhou2021ibot}) between the masked student tokens and the corresponding visible teacher tokens. Fig.~\ref{fig:pipeline} illustrates this setup.

We do this self-supervised pre-training using 1 million face crops and hand crops (obtained as described in Section (\S \ref{sec:data_preprocessing})) independently. We use a base learning rate of $2\times10^{-4}$ and batch size per GPU of 128 on 8 A6000Ada GPUs (i.e., an effective batch size of 1024). To account for our relatively small dataset, we follow the recommendation of \citet{roth2024lowresource} and adjust the number of iterations per pseudo-epoch and the number of pseudo-epochs, resulting in 5 effective epochs in total. In the KoLeoLoss we change the hyperparameter $\epsilon$ from $10^{-8}$ to $10^{-4}$ to avoid infinite loss values.

The resulting teacher networks are used as our face encoder (DINOv2-F) and hand encoder (DINOv2-H) for the next stage, which is supervised training. We note that for the face, we found that blurring the whole face directly hurts performance. However, a modified approach of first greying areas except the eyes and mouth, and then blurring the remaining image, performs similarly to training on the whole face crop.  We therefore use this strategy, which aims to balance between keeping important facial attributes while also having some privacy-awareness.

\subsection{Supervised Sign Language Translation}
\label{sec:supervised_sign_language_translation}

Given a video of $T$ frames with an associated written language translation, we first obtain channel crops as described in (\S \ref{sec:data_preprocessing}). Each crop is passed to the relevant frozen encoder, which is obtained as described in (\S \ref{sec:self-supervised_sign_components_pre-training}). This results in 3 $T \times 384$ matrices for the face, left hand and right hand. These matrices are projected to $T \times 256$ via stream-specific linear layers. The normalized body pose vectors are transformed to a higher dimensionality, from $T \times 14$ to $T \times128$ (via a linear layer trained from scratch). All four feature streams (face, left hand, right hand, and body pose) are then temporally concatenated and projected to $T \times 768$ (the input size of the T5 model) via another linear layer, also trained from scratch.

To summarize: In stage 1, we independently pre-train two (face- and hand-specific) image-to-vector feature extractors. In stage 2, we jointly train a human pose feature linear transformation layer and a single linear layer transformation for the concatenated four-stream features, and we fine-tune the T5 model for translation. Both stages are shown in~\Cref{fig:overview}.

\section{Experiments}
\label{sec:experiments}

\begin{table*}[ht!]
\centering
\vspace{8pt}
\fontsize{10}{12}\selectfont
\sisetup{table-format = 3.2}
\resizebox{1.0\textwidth}{!}{
\begin{tabular}{@{}lrrrrrr@{}}
\toprule
\textsc{Method} & \multicolumn{1}{l}{\textsc{BLEU-1}} & \textsc{BLEU-2} & \textsc{BLEU-3} & \textsc{BLEU} & \multicolumn{1}{l}{GPU-Hrs} & \multicolumn{1}{l}{PT \% of Frames} \\
\midrule
\addlinespace[0.3em]
\rowcol\multicolumn{7}{c}{\textbf{Supervised training Schedule: H2S}} \\
\addlinespace[0.2em]

\citet{lin-etal-2023-gloss}  & \num{14.9} & \num{7.3} & \num{3.9} & \num{2.2} & --- & --- \\

\citet{slt-how2sign-wicv2023} & \num{34.0} & \num{19.3} & \num{12.2} & \num{8.0} & --- & ---\\

\citet{uthus2023youtube} & \num{15.0} & \num{5.1} & \num{2.3} & \num{1.2} & {---} & {---}\\

\ssvp $_ {800}^{\text{YT}(50)}$~\cite{rust2024towards} & \num{38.1} & \num{23.7} & \num{16.3} & \num{11.7} & {$18535^*$} & $50\hspace{10pt}$\\

SignMusketeers $_{5}^{\text{YT} (1.2) }$ (Ours) & \num{18.8} & \num{8.1} & \num{4.2} & \num{2.4} & 592 & $1.2\hspace{10pt}$\\
\addlinespace[0.3em]
\rowcol\multicolumn{7}{c}{\textbf{Supervised training Schedule: YT}} \\
\addlinespace[0.2em]

\citet{uthus2023youtube} & \num{20.9} & \num{10.4} & \num{6.1} & \num{4.0} & --- &---\\

\ssvp $_ {800}^{\text{YT}(50)}$~\cite{rust2024towards} & \num{29.2} & \num{16.6} & \num{10.7} & \num{7.1} & $18729^*$ &$50\hspace{10pt}$\\

SignMusketeers $_{5}^{\text{YT} (1.2) }$ (Ours) & \num{26.3} & \num{13.8} & \num{8.2} & \num{5.2} & 864 & $1.2\hspace{10pt}$\\

\addlinespace[0.3em]
\rowcol\multicolumn{7}{c}{\textbf{*Supervised training Schedule: YT $+$ H2S}} \\
\addlinespace[0.2em]

\citet{uthus2023youtube} & \num{36.3} & \num{23.0} & \num{16.1} & \num{11.9} & --- &---\\

\ssvp $_ {800}^{\text{YT}(50)}$~\cite{rust2024towards} & \num{41.6} & \num{27.2} & \num{19.3} & \num{14.3} & $18768^*$ &  $50\hspace{10pt}$\\

\ssvp $_{100}^{\text{YT$+$H2S}(50)}$~\cite{rust2024towards} & --- & --- & --- & \num{12.5} & $2754^*$ & $50\hspace{10pt}$\\

\ssvp-\textsc{LSP} $_{600 \rightarrow 200}^{\text{YT} + \text{H2S}}$ ~\cite{rust2024towards}& \num{43.2} & \num{28.8} & \num{20.8} & \num{15.5} & $32912^*$ & $50\hspace{10pt}$ \\

\addlinespace[0.3em]
\rowcol\multicolumn{7}{c}{\textbf{Supervised training Schedule: YT $\rightarrow$ H2S}} \\
\addlinespace[0.2em]

\citet{uthus2023youtube} & \num{37.8} & \num{24.1} & \num{16.9} & \num{12.4} & --- &---\\

\ssvp $_ {800}^{\text{YT}(50)}$~\cite{rust2024towards} & \num{41.9} & \num{27.7} & \num{19.8} & \num{14.7} & $18768^*$ & ---\\

SignMusketeers $_{5}^{\text{YT} (1.2) }$ (Ours) & \num{41.5} & \num{27.2} & \num{19.3} & \num{14.3} & 880 & $1.2\hspace{10pt}$\\

\addlinespace[0.2em]
\bottomrule
\end{tabular}
}
\caption{Quantitative results on How2Sign. 
GPU-Hrs = GPU hours used during the entire training stage. PT \% of frames = Percentage of YouTube-ASL frames used in the self-supervised pre-training stage. $^*$Adjusted by throughput ratio reported at \url{https://lambdalabs.com/gpu-benchmarks}. *We did not include YT + H2S experiments as this setting does not reflect the common paradigm of pre-training on a large dataset and fine-tuning on a different smaller dataset. The YT → H2S paradigm better represents adaptability to real-world use cases.}
\label{tab:main_results}
\end{table*}

As in other work~\cite{uthus2023youtube, rust2024towards}, we use the YouTube-ASL and How2Sign \cite{duarte2021how2sign} datasets. YouTube-ASL contains roughly 600,000 clips, or roughly 700 hours, of ASL video with weakly aligned English text translations.\footnote{Although \citet{uthus2023youtube} report a total of 610,193 clips in YouTube-ASL, we were only able to retrieve 601,995 clips, presumably because some clips have been deleted between the time of the dataset's creation and our retrieval.}  How2Sign consists of 31,128 / 1,741 / 2,322 clips for the training / validation / test sets.

For the self-supervised training of DINOv2-Hand and DINOv2-Face, due to computational constraints we limit ourselves to 1 million random face crops and 1 million random hand crops from YouTube-ASL. We note that the state-of-the-art method proposed by \citet{rust2024towards} sees about 50 million frames during pre-training. 

In line with previous studies, we employ the following training schedules for the supervised (translation) stage, using a stride of 2 for every video clip:

\vspace{.1in}
  \noindent\textbf{H2S}: Supervised training exclusively on the How2Sign dataset, without using the YouTube-ASL dataset.

  \vspace{.1in}
  \noindent\textbf{YT}: Supervised training solely on YouTube-ASL, and evaluation on How2Sign in a zero-shot setting.

  \vspace{.1in}
  \noindent\textbf{YT$\to$H2S}: Supervised training on YouTube-ASL, followed by supervised fine-tuning on How2Sign.

\vspace{.1in}
  Note that \citet{uthus2023youtube}
 and \citet{rust2024towards} include an additional training schedule, YT $+$ H2S, which involves training on a mixture of YT and H2S.

For the supervised training stage, similarly to other work \cite{uthus2023youtube, rust2024towards}, we initiliaze our T5 from a T5.1.1-Base pre-trained checkpoint. We use a batch size of 128 (16 per GPU running on 8 GPUs), with other hyper-parameters identical to \citet{rust2024towards}. Additionally, when further fine-tuning on How2Sign after training on YouTube-ASL, we perform an extra 5,000 steps of fine-tuning.

 \subsection{Comparison to prior work}\label{sec:results}

Table~\ref{tab:main_results} compares our models to prior results on the How2Sign ASL-English translation task, including two approaches that train on a combination of YouTube-ASL and How2Sign. First, we observe that our method consistently outperforms the approach of \citet{uthus2023youtube} across various metrics. For example, we improve BLEU by 1.9 points when using the YT→H2S training schedule. This improvement provides evidence for the benefits of using learned features over pose  estimator features/coordinates. When using the H2S-only supervised training schedule, our scores are 1.2 BLEU scores above the ones of \citet{uthus2023youtube} on this same training schedule (1.2 BLEU vs.~2.4 BLEU).

Comparing with the SSVP-SLT approach of~\citet{rust2024towards}, in the most restrictive H2S-only schedule, our method is far behind (by 9 BLEU points). However, in the best-performing schedule (YT→H2S), the gap between the two approaches reduces significantly, and our performance is just 0.4 BLEU points below that of~\citet{rust2024towards}.\footnote{It is important to note that the authors of ~\citet{rust2024towards} report an even higher BLEU score of 15.5 for a method that trains an additional CLIP~\cite{radford2021learning} model on English text, on the union of YouTube-ASL and How2Sign. We include this result in \Cref{fig:bubble} (top right). Such techniques might also improve our performance, but we are unable to do the experiment due to computation constraints.} We note that the same trend holds for the method of \citet{uthus2023youtube} when compared to SSVP-SLT. On H2S alone, there is a substantial gap between the two methods (+10.5 BLEU points), but on YT→H2S, the gap reduces to +2.4 BLEU points.

Human pose estimation, as used in the method of \citet{uthus2023youtube}, is inherently multi-stream since it returns vectors (coordinates) describing multiple body parts, which are later concatenated. We suspect that both of the multi-stream approaches
do not perform well with small datasets in supervised training. We hypothesize that this might be because the features they return are entirely frame-level features and do not \textit{yet} contain any information about how these frame-level features relate to each other. In contrast, SSVP-SLT pre-trained features already have some information about how frames relate to each other.

Nevertheless, with a larger dataset during the supervised training stage, the lead of SSVP-SLT diminishes drastically compared to multi-stream approaches like \citet{uthus2023youtube} and our SignMusketeers method. This suggests that as the amount of labeled training data increases, the advantage of pre-trained features that capture some temporal relationships becomes less pronounced, and multi-stream approaches can bridge the performance gap.

We suspect that the 0.4 BLEU difference between our result and that of SSVP-SLT may be attributable not only to the model architecture, but also to other significant factors that vary between the two approaches.
One key difference is the amount of data used during pre-training. Our method uses a sample of only 1.2\% of the YouTube-ASL frames, while SSVP-SLT uses 50\% of the YouTube-ASL frames (every video at a stride of 2). Additionally, due to computational constraints, we pre-train our model for only 5 epochs, whereas SSVP-SLT pre-trains for 800 epochs. As another comparison, at epoch 100, SSVP-SLT achieves a BLEU score of 12.5 (see \Cref{tab:main_results} {\tiny \ssvp $_{100}^{\text{YT$+$H2S}(50)}$} and \Cref{fig:bubble}). In addition, we note that SSVP-SLT is pre-trained on both YouTube-ASL and How2Sign---the target domain dataset---while our model is pre-trained only on YouTube-ASL. Fig.~\ref{fig:bubble} shows the performance-resources tradeoff for multiple models, showing that our approach (top left, in blue) surpasses SSVP-SLT by 1.7 BLEU points while training for 20 times fewer pre-training epochs and without utilizing the target domain dataset during pre-training.

These observations suggest that our method, despite using significantly less pre-training data and fewer pre-training epochs, can achieve competitive performance compared to the state of the art. Further investigation into the impact of pre-training data size and the number of pre-training epochs on the final performance could provide valuable insights into the efficiency and scalability of sign language translation models.

We perform ablation studies to investigate three key components of the model design, as shown in \Cref{sec:ablation}. First, pre-training DINOv2 specifically on hand and face crops proves beneficial, improving performance. Second, the multi-stream approach demonstrates significant advantages over using just the original frames, significantly boosting the scores. Finally, adding raw frames as a fifth stream alongside the existing features (perhaps suprisingly) degrades performance. See \Cref{sec:ablation} for more details.

\section{Conclusion}
\label{sec:conclusion}
SignMusketeers is a data- and compute-efficient method for sign language translation at scale. It uses a multi-stream encoding scheme that focuses on important parts of a signing video (facial expressions, hands, and body pose) and requires only individual frames during self-supervised pre-training, in contrast to prior work on pre-training that requires long duration videos. SignMusketeers achieves competitive performance with roughly $40\times$ less data and $50\times$ less computation than prior work. %

\paragraph{Limitations}

This research has several key limitations. The most significant is that despite matching or exceeding previous studies' results, the overall performance remains inadequate. Therefore, machine translation cannot yet reliably replace human sign language interpreters across diverse contexts. Another limitation stems from the much smaller training datasets compared to those typically used for developing self-supervised models in the speech and text domains. Additionally, since this study focuses solely on American Sign Language, we cannot determine whether our approach would work effectively for other sign languages, even though different sign languages share common features and communication methods. Also, in this work we focus only on the sign language translation task from video to text. It will be important to adapt this approach to the opposite direction (text to video) and other sign language processing tasks such as isolated sign language recognition, continuous sign language recognition, and fingerspelling detection. To overcome these challenges, future research should focus on building larger datasets and incorporating multiple sign languages into the training process.

\paragraph{Acknowledgment}
We are grateful to Anand Bhattad, Ankita Pasad, Ju-Chieh Chou, and Chung-Ming Chien for their valuable suggestions throughout this project.

\bibliography{custom}

\begin{thebibliography}{61}
\providecommand{\natexlab}[1]{#1}

\bibitem[{Albanie et~al.(2021)Albanie, Varol, Momeni, Bull, Afouras, Chowdhury, Fox, Woll, Cooper, McParland et~al.}]{Albanie2021bobsl}
Samuel Albanie, Gül Varol, Liliane Momeni, Hannah Bull, Triantafyllos Afouras, Himel Chowdhury, Neil Fox, Bencie Woll, Rob Cooper, Andrew McParland, et~al. 2021.
\newblock {BBC-Oxford British Sign Language} dataset.
\newblock \emph{arXiv preprint arXiv:2111.03635}.

\bibitem[{Boháček and Hrúz(2022)}]{bohavcek2022sign}
Matyáš Boháček and Marek Hrúz. 2022.
\newblock Sign pose-based transformer for word-level sign language recognition.
\newblock In \emph{Proc. Winter Conference on Applications of Computer Vision (WACV)}.

\bibitem[{Bragg et~al.(2021)Bragg, Caselli, Gallagher, Goldberg, Oka, and Thies}]{bragg2021asl}
Danielle Bragg, Naomi Caselli, John~W Gallagher, Miriam Goldberg, Courtney~J Oka, and William Thies. 2021.
\newblock {ASL} sea battle: gamifying sign language data collection.
\newblock In \emph{Proc. CHI}.

\bibitem[{Bragg et~al.(2019)Bragg, Koller, Bellard, Berke, Boudreault, Braffort, Caselli, Huenerfauth, Kacorri, Verhoef et~al.}]{bragg2019sign}
Danielle Bragg, Oscar Koller, Mary Bellard, Larwan Berke, Patrick Boudreault, Annelies Braffort, Naomi Caselli, Matt Huenerfauth, Hernisa Kacorri, Tessa Verhoef, et~al. 2019.
\newblock Sign language recognition, generation, and translation: An interdisciplinary perspective.
\newblock In \emph{Proc. SIGACCESS}.

\bibitem[{Brentari(1998)}]{brentari1998prosodic}
Diane Brentari. 1998.
\newblock \emph{A Prosodic Model of Sign Language Phonology}.
\newblock MIT Press.

\bibitem[{Brentari(2019)}]{brentari2019sign}
Diane Brentari. 2019.
\newblock \emph{Sign Language Phonology}.
\newblock Cambridge University Press.

\bibitem[{Cai et~al.(2023)Cai, Ghosh, Stefanov, Dhall, Cai, Rezatofighi, Haffari, and Hayat}]{cai2023marlin}
Zhixi Cai, Shreya Ghosh, Kalin Stefanov, Abhinav Dhall, Jianfei Cai, Hamid Rezatofighi, Reza Haffari, and Munawar Hayat. 2023.
\newblock Marlin: Masked autoencoder for facial video representation learning.
\newblock In \emph{Proc. CVPR}.

\bibitem[{Camgoz et~al.(2018)Camgoz, Hadfield, Koller, Ney, and Bowden}]{camgoz2018neural}
Necati~Cihan Camgoz, Simon Hadfield, Oscar Koller, Hermann Ney, and Richard Bowden. 2018.
\newblock Neural sign language translation.
\newblock In \emph{Proc. CVPR}.

\bibitem[{Camgoz et~al.(2020)Camgoz, Koller, Hadfield, and Bowden}]{camgoz2020multi}
Necati~Cihan Camgoz, Oscar Koller, Simon Hadfield, and Richard Bowden. 2020.
\newblock Multi-channel transformers for multi-articulatory sign language translation.
\newblock In \emph{ECCV Workshops}.

\bibitem[{Cao et~al.(2017)Cao, Simon, Wei, and Sheikh}]{cao2017realtime}
Zhe Cao, Tomas Simon, Shih-En Wei, and Yaser Sheikh. 2017.
\newblock Realtime multi-person 2d pose estimation using part affinity fields.
\newblock In \emph{Proc. CVPR}.

\bibitem[{Caron et~al.(2020)Caron, Misra, Mairal, Goyal, Bojanowski, and Joulin}]{caron2020unsupervised}
Mathilde Caron, Ishan Misra, Julien Mairal, Priya Goyal, Piotr Bojanowski, and Armand Joulin. 2020.
\newblock Unsupervised learning of visual features by contrasting cluster assignments.
\newblock \emph{Proc. NeurIPS}.

\bibitem[{Contributors(2020)}]{mmpose2020}
MMPose Contributors. 2020.
\newblock Openmmlab pose estimation toolbox and benchmark.
\newblock \url{https://github.com/open-mmlab/mmpose}.

\bibitem[{da~Silva et~al.(2020)da~Silva, Costa, Kumada, De~Martino, and Florentino}]{da2020recognition}
Emely~Pujólli da~Silva, Paula Dornhofer~Paro Costa, Kate Mamhy~Oliveira Kumada, José~Mario De~Martino, and Gabriela~Araújo Florentino. 2020.
\newblock Recognition of affective and grammatical facial expressions: A study for brazilian sign language.
\newblock In \emph{ECCV Workshops}.

\bibitem[{Darcet et~al.(2024)Darcet, Oquab, Mairal, and Bojanowski}]{darcet2024vision}
Timothée Darcet, Maxime Oquab, Julien Mairal, and Piotr Bojanowski. 2024.
\newblock Vision transformers need registers.
\newblock In \emph{The Twelfth International Conference on Learning Representations}.

\bibitem[{De~Coster et~al.(2023)De~Coster, Rushe, Holmes, Ventresque, and Dambre}]{de2023towards}
Mathieu De~Coster, Ellen Rushe, Ruth Holmes, Anthony Ventresque, and Joni Dambre. 2023.
\newblock Towards the extraction of robust sign embeddings for low resource sign language recognition.
\newblock \emph{arXiv preprint arXiv:2306.17558}.

\bibitem[{Desai et~al.(2024)Desai, De~Meulder, Hochgesang, Kocab, and Lu}]{desai2024systemic}
Aashaka Desai, Maartje De~Meulder, Julie~A Hochgesang, Annemarie Kocab, and Alex~X Lu. 2024.
\newblock Systemic biases in sign language ai research: A deaf-led call to reevaluate research agendas.
\newblock \emph{arXiv preprint arXiv:2403.02563}.

\bibitem[{Duarte et~al.(2021)Duarte, Palaskar, Ventura, Ghadiyaram, DeHaan, Metze, Torres, and Giro-i Nieto}]{duarte2021how2sign}
Amanda Duarte, Shruti Palaskar, Lucas Ventura, Deepti Ghadiyaram, Kenneth DeHaan, Florian Metze, Jordi Torres, and Xavier Giro-i Nieto. 2021.
\newblock {How2Sign}: A large-scale multimodal dataset for continuous {American Sign Language}.
\newblock In \emph{Proc. CVPR}.

\bibitem[{Dumas(1894)}]{dumas1894works}
Alexandre Dumas. 1894.
\newblock \emph{The Works of Alexandre Dumas: Three Musketeers}.
\newblock Estes and Lauriat.

\bibitem[{Fox et~al.(2023)Fox, Woll, and Cormier}]{fox2023best}
Neil Fox, Bencie Woll, and Kearsy Cormier. 2023.
\newblock Best practices for sign language technology research.
\newblock \emph{Universal Access in the Information Society}.

\bibitem[{Gao and Patras(2024)}]{gao2024self}
Zheng Gao and Ioannis Patras. 2024.
\newblock Self-supervised facial representation learning with facial region awareness.
\newblock \emph{arXiv preprint arXiv:2403.02138}.

\bibitem[{Grill et~al.(2020)Grill, Strub, Altché, Tallec, Richemond, Buchatskaya, Doersch, Avila~Pires, Guo, Gheshlaghi~Azar et~al.}]{grill2020bootstrap}
Jean-Bastien Grill, Florian Strub, Florent Altché, Corentin Tallec, Pierre Richemond, Elena Buchatskaya, Carl Doersch, Bernardo Avila~Pires, Zhaohan Guo, Mohammad Gheshlaghi~Azar, et~al. 2020.
\newblock Bootstrap your own latent: A new approach to self-supervised learning.
\newblock \emph{Proc. NeurIPS}.

\bibitem[{Gueuwou et~al.(2023)Gueuwou, Siake, Leong, and Müller}]{gueuwou-etal-2023-jwsign}
Shester Gueuwou, Sophie Siake, Colin Leong, and Mathias Müller. 2023.
\newblock {JWSign}: A highly multilingual corpus of bible translations for more diversity in sign language processing.
\newblock In \emph{Findings of the Association for Computational Linguistics: EMNLP}.

\bibitem[{Hachiuma et~al.(2023)Hachiuma, Sato, and Sekii}]{hachiuma2023unified}
Ryo Hachiuma, Fumiaki Sato, and Taiki Sekii. 2023.
\newblock Unified keypoint-based action recognition framework via structured keypoint pooling.
\newblock In \emph{Proc. CVPR}.

\bibitem[{Holden et~al.(2005)Holden, Lee, and Owens}]{holden2005australian}
Eun-Jung Holden, Gareth Lee, and Robyn Owens. 2005.
\newblock Australian sign language recognition.
\newblock \emph{Machine Vision and Applications}.

\bibitem[{Koller et~al.(2016)Koller, Ney, and Bowden}]{koller2016deep}
Oscar Koller, Hermann Ney, and Richard Bowden. 2016.
\newblock Deep hand: How to train a cnn on 1 million hand images when your data is continuous and weakly labelled.
\newblock In \emph{Proc. CVPR}.

\bibitem[{Kuznetsova and Kimmelman(2024)}]{kuznetsova2024testing}
Anna Kuznetsova and Vadim Kimmelman. 2024.
\newblock Testing {MediaPipe Holistic} for linguistic analysis of nonmanual markers in sign languages.
\newblock \emph{arXiv preprint arXiv:2403.10367}.

\bibitem[{Liddell and Johnson(1989)}]{liddell1989american}
Scott~K Liddell and Robert~E Johnson. 1989.
\newblock {American Sign Language}: The phonological base.
\newblock \emph{Sign Language Studies}.

\bibitem[{Lin et~al.(2023)Lin, Wang, Zhu, Sun, Zhang, and Yang}]{lin-etal-2023-gloss}
Kezhou Lin, Xiaohan Wang, Linchao Zhu, Ke~Sun, Bang Zhang, and Yi~Yang. 2023.
\newblock Gloss-free end-to-end sign language translation.
\newblock In \emph{Proc. ACL}.

\bibitem[{Lugaresi et~al.(2019)Lugaresi, Tang, Nash, McClanahan, Uboweja, Hays, Zhang, Chang, Yong, Lee et~al.}]{lugaresi2019mediapipe}
Camillo Lugaresi, Jiuqiang Tang, Hadon Nash, Chris McClanahan, Esha Uboweja, Michael Hays, Fan Zhang, Chuo-Ling Chang, Ming~Guang Yong, Juhyun Lee, et~al. 2019.
\newblock Mediapipe: A framework for building perception pipelines.
\newblock \emph{arXiv preprint arXiv:1906.08172}.

\bibitem[{Moryossef et~al.(2021)Moryossef, Tsochantaridis, Dinn, Camgoz, Bowden, Jiang, Rios, Müller, and Ebling}]{moryossef2021evaluating}
Amit Moryossef, Ioannis Tsochantaridis, Joe Dinn, Necati~Cihan Camgoz, Richard Bowden, Tao Jiang, Annette Rios, Mathias Müller, and Sarah Ebling. 2021.
\newblock Evaluating the immediate applicability of pose estimation for sign language recognition.
\newblock In \emph{Proc. CVPR}.

\bibitem[{Müller et~al.(2023)Müller, Alikhani, Avramidis, Bowden, Braffort, Camgoz, Ebling, España-Bonet, Göhring, Grundkiewicz et~al.}]{muller-etal-2023-findings}
Mathias Müller, Malihe Alikhani, Eleftherios Avramidis, Richard Bowden, Annelies Braffort, Necati~Cihan Camgoz, Sarah Ebling, Cristina España-Bonet, Anne Göhring, Roman Grundkiewicz, et~al. 2023.
\newblock Findings of the second wmt shared task on sign language translation (wmt-slt23).
\newblock In \emph{Proc. Conference on Machine Translation (WMT)}.

\bibitem[{Müller et~al.(2022)Müller, Ebling, Avramidis, Battisti, Berger, Bowden, Braffort, Camgoz, España-Bonet, Grundkiewicz et~al.}]{muller-etal-2022-findings}
Mathias Müller, Sarah Ebling, Eleftherios Avramidis, Alessia Battisti, Michèle Berger, Richard Bowden, Annelies Braffort, Necati~Cihan Camgoz, Cristina España-Bonet, Roman Grundkiewicz, et~al. 2022.
\newblock Findings of the first wmt shared task on sign language translation (wmt-slt22).
\newblock In \emph{Proc. Conference on Machine Translation (WMT)}.

\bibitem[{Oh et~al.(2016)Oh, Benenson, Fritz, and Schiele}]{oh2016faceless}
Seong~Joon Oh, Rodrigo Benenson, Mario Fritz, and Bernt Schiele. 2016.
\newblock Faceless person recognition: Privacy implications in social media.
\newblock In \emph{Proc. ECCV}.

\bibitem[{Oquab et~al.(2023)Oquab, Darcet, Moutakanni, Vo, Szafraniec, Khalidov, Fernandez, Haziza, Massa, El-Nouby et~al.}]{oquab2024dinov}
Maxime Oquab, Timothée Darcet, Théo Moutakanni, Huy~V Vo, Marc Szafraniec, Vasil Khalidov, Pierre Fernandez, Daniel Haziza, Francisco Massa, Alaaeldin El-Nouby, et~al. 2023.
\newblock {DINOv2}: Learning robust visual features without supervision.
\newblock \emph{Trans. Machine Learning Research}.

\bibitem[{Pfau et~al.(2010)Pfau, Quer et~al.}]{Pfau2010NonmanualsTG}
Roland Pfau, Josep Quer, et~al. 2010.
\newblock \emph{Nonmanuals: Their Grammatical and Prosodic Roles}.
\newblock Cambridge University Press.

\bibitem[{Post(2018)}]{post-2018-call}
Matt Post. 2018.
\newblock A call for clarity in reporting {BLEU} scores.
\newblock In \emph{Proc. Conference on Machine Translation (WMT)}.

\bibitem[{Pu et~al.(2016)Pu, Zhou, and Li}]{pu2016sign}
Junfu Pu, Wengang Zhou, and Houqiang Li. 2016.
\newblock Sign language recognition with multi-modal features.
\newblock In \emph{Advances in Multimedia Information Processing--PCM 2016: 17th Pacific-Rim Conference on Multimedia}.

\bibitem[{Radford et~al.(2021)Radford, Kim, Hallacy, Ramesh, Goh, Agarwal, Sastry, Askell, Mishkin, Clark et~al.}]{radford2021learning}
Alec Radford, Jong~Wook Kim, Chris Hallacy, Aditya Ramesh, Gabriel Goh, Sandhini Agarwal, Girish Sastry, Amanda Askell, Pamela Mishkin, Jack Clark, et~al. 2021.
\newblock Learning transferable visual models from natural language supervision.
\newblock In \emph{International Conference on Machine Learning}.

\bibitem[{Raffel et~al.(2020)Raffel, Shazeer, Roberts, Lee, Narang, Matena, Zhou, Li, and Liu}]{raffel2020exploring}
Colin Raffel, Noam Shazeer, Adam Roberts, Katherine Lee, Sharan Narang, Michael Matena, Yanqi Zhou, Wei Li, and Peter~J Liu. 2020.
\newblock Exploring the limits of transfer learning with a unified text-to-text transformer.
\newblock \emph{Journal of Machine Learning Research}.

\bibitem[{Roth et~al.(2024)Roth, Koch, Wagner, Schnabel, Marr, and Peng}]{roth2024lowresource}
Benedikt Roth, Valentin Koch, Sophia~J Wagner, Julia~A Schnabel, Carsten Marr, and Tingying Peng. 2024.
\newblock Low-resource finetuning of foundation models beats state-of-the-art in histopathology.

\bibitem[{Rust et~al.(2024)Rust, Shi, Wang, Camgoz, and Maillard}]{rust2024towards}
Phillip Rust, Bowen Shi, Skyler Wang, Necati~Cihan Camgoz, and Jean Maillard. 2024.
\newblock Towards privacy-aware sign language translation at scale.
\newblock In \emph{Proc. ACL}.

\bibitem[{Ryali et~al.(2023)Ryali, Hu, Bolya, Wei, Fan, Huang, Aggarwal, Chowdhury, Poursaeed, Hoffman et~al.}]{ryali2023hiera}
Chaitanya Ryali, Yuan-Ting Hu, Daniel Bolya, Chen Wei, Haoqi Fan, Po-Yao Huang, Vaibhav Aggarwal, Arkabandhu Chowdhury, Omid Poursaeed, Judy Hoffman, et~al. 2023.
\newblock Hiera: A hierarchical vision transformer without the bells-and-whistles.
\newblock In \emph{Proc. ICML}.

\bibitem[{Sandler and Lillo-Martin(2006)}]{sandler2006sign}
Wendy Sandler and Diane~Carolyn Lillo-Martin. 2006.
\newblock \emph{Sign Language and Linguistic Universals}.
\newblock Cambridge University Press.

\bibitem[{Sandoval-Castaneda et~al.(2023)Sandoval-Castaneda, Li, Brentari, Livescu, and Shakhnarovich}]{sandoval2023self}
Marcelo Sandoval-Castaneda, Yanhong Li, Diane Brentari, Karen Livescu, and Gregory Shakhnarovich. 2023.
\newblock Self-supervised video transformers for isolated sign language recognition.
\newblock \emph{arXiv preprint arXiv:2309.02450}.

\bibitem[{Santhalingam et~al.(2020)Santhalingam, Pathak, Rangwala, Košecká et~al.}]{santhalingam2020finehand}
Panneer~Selvam Santhalingam, Parth Pathak, Huzefa Rangwala, Jana Košecká, et~al. 2020.
\newblock {FineHand}: Learning hand shapes for {American Sign Language} recognition.
\newblock In \emph{IEEE International Conference on Automatic Face and Gesture Recognition (FG)}.

\bibitem[{Saunders et~al.(2020)Saunders, Camgoz, and Bowden}]{saunders2020adversarial}
Ben Saunders, Necati~Cihan Camgoz, and Richard Bowden. 2020.
\newblock Adversarial training for multi-channel sign language production.
\newblock \emph{arXiv preprint arXiv:2008.12405}.

\bibitem[{Shi et~al.(2022)Shi, Brentari, Shakhnarovich, and Livescu}]{shi-etal-2022-open}
Bowen Shi, Diane Brentari, Gregory Shakhnarovich, and Karen Livescu. 2022.
\newblock Open-domain sign language translation learned from online video.
\newblock In \emph{Proc. EMNLP}.

\bibitem[{Sutton-Spence and Woll(1999)}]{Sutton-Spence_Woll_1999}
Rachel Sutton-Spence and Bencie Woll. 1999.
\newblock \emph{The Linguistics of {British Sign Language}: An Introduction}.
\newblock Cambridge University Press.

\bibitem[{Tarrés et~al.(2023)Tarrés, Gállego, Duarte, Torres, and Giró-i Nieto}]{slt-how2sign-wicv2023}
Laia Tarrés, Gerard~I Gállego, Amanda Duarte, Jordi Torres, and Xavier Giró-i Nieto. 2023.
\newblock Sign language translation from instructional videos.
\newblock In \emph{Proc. CVPR}.

\bibitem[{Tornay et~al.(2020)Tornay, Camgoz, Bowden, and Magimai~Doss}]{tornay2020phonology}
Sandrine Tornay, Necati~Cihan Camgoz, Richard Bowden, and Mathew Magimai~Doss. 2020.
\newblock A phonology-based approach for isolated sign production assessment in sign language.
\newblock In \emph{Companion Publication of the 2020 International Conference on Multimodal Interaction}.

\bibitem[{Uthus et~al.(2023)Uthus, Tanzer, and Georg}]{uthus2023youtube}
Dave Uthus, Garrett Tanzer, and Manfred Georg. 2023.
\newblock {YouTube-ASL}: A large-scale, open-domain {American Sign Language-English} parallel corpus.
\newblock In \emph{Proc. NeurIPS}.

\bibitem[{Varol et~al.(2021)Varol, Momeni, Albanie, Afouras, and Zisserman}]{varol2021read}
Gul Varol, Liliane Momeni, Samuel Albanie, Triantafyllos Afouras, and Andrew Zisserman. 2021.
\newblock Read and attend: Temporal localisation in sign language videos.
\newblock In \emph{Proc. CVPR}.

\bibitem[{Vogler and Metaxas(2001)}]{vogler2001framework}
Christian Vogler and Dimitris Metaxas. 2001.
\newblock A framework for recognizing the simultaneous aspects of {American Sign Language}.
\newblock \emph{Computer Vision and Image Understanding}.

\bibitem[{Wong et~al.(2024)Wong, Camgoz, and Bowden}]{wong2024signgpt}
Ryan Wong, Necati~Cihan Camgoz, and Richard Bowden. 2024.
\newblock {Sign2GPT}: Leveraging large language models for gloss-free sign language translation.
\newblock In \emph{Proc. ICLR}.

\bibitem[{Yan et~al.(2023)Yan, Liu, Wei, Li, Li, and Lin}]{yan2023skeletonmae}
Hong Yan, Yang Liu, Yushen Wei, Zhen Li, Guanbin Li, and Liang Lin. 2023.
\newblock Skeletonmae: Graph-based masked autoencoder for skeleton sequence pre-training.
\newblock In \emph{Proc. CVPR}.

\bibitem[{Yin et~al.(2021)Yin, Moryossef, Hochgesang, Goldberg, and Alikhani}]{yin-etal-2021-including}
Kayo Yin, Amit Moryossef, Julie Hochgesang, Yoav Goldberg, and Malihe Alikhani. 2021.
\newblock Including signed languages in natural language processing.
\newblock In \emph{Proc. ACL-IJCNLP}.

\bibitem[{Zafrulla et~al.(2010)Zafrulla, Brashear, Yin, Presti, Starner, and Hamilton}]{zafrulla2010american}
Zahoor Zafrulla, Helene Brashear, Pei Yin, Peter Presti, Thad Starner, and Harley Hamilton. 2010.
\newblock {American Sign Language} phrase verification in an educational game for deaf children.
\newblock In \emph{Proc. ICPR}.

\bibitem[{Zhou et~al.(2021{\natexlab{a}})Zhou, Zhou, Qi, Pu, and Li}]{zhou2021improving}
Hao Zhou, Wengang Zhou, Weizhen Qi, Junfu Pu, and Houqiang Li. 2021{\natexlab{a}}.
\newblock Improving sign language translation with monolingual data by sign back-translation.
\newblock In \emph{Proc. CVPR}.

\bibitem[{Zhou et~al.(2021{\natexlab{b}})Zhou, Zhou, Zhou, and Li}]{zhou2021spatial}
Hao Zhou, Wengang Zhou, Yun Zhou, and Houqiang Li. 2021{\natexlab{b}}.
\newblock Spatial-temporal multi-cue network for sign language recognition and translation.
\newblock \emph{IEEE Trans. Multimedia}, 24:768--779.

\bibitem[{Zhou et~al.(2022)Zhou, Wei, Wang, Shen, Xie, Yuille, and Kong}]{zhou2021ibot}
Jinghao Zhou, Chen Wei, Huiyu Wang, Wei Shen, Cihang Xie, Alan Yuille, and Tao Kong. 2022.
\newblock {iBOT}: Image {BERT} pre-training with online tokenizer.
\newblock \emph{Proc. ICLR}.

\bibitem[{Zimmermann et~al.(2021)Zimmermann, Argus, and Brox}]{zimmermann2021contrastive}
Christian Zimmermann, Max Argus, and Thomas Brox. 2021.
\newblock Contrastive representation learning for hand shape estimation.
\newblock In \emph{DAGM German Conference on Pattern Recognition}.

\end{thebibliography}
\appendix

\section{Ablations}
\label{sec:ablation}

\begin{table}[t!]
    \centering
    \resizebox{\linewidth}{!}{%
    \begin{tabular}{l r r r r }
        \toprule
         & \textsc{BLEU-1} & \textsc{BLEU-2} & \textsc{BLEU-3} & \textsc{BLEU}\\
        \midrule 
        Uncrop $+$ orig. & 23.5 & 12.4 & 7.6 & 4.9\\
        Crop $+$ orig. & 36.5 & 23.1 & 15.9 & 11.3\\
        Crop $+$ pre-tr.  & 36.7 & 23.8 & 16.7 & 12.2\\
        \hspace{.01in} $+$ global frame & & & & \\
        \midrule
        Crop $+$ pre-tr.  & {\bf 41.5} & {\bf 27.2} & {\bf 19.3} &{\bf 14.3}\\
        \hspace{.01in} (Ours) & & & & \\
        \bottomrule
    \end{tabular}
    }
        \caption{Ablation studies of our design choices. All models are pre-trained on YT in stage1 and fine-tuned with YT $\to$ H2S schedule for stage2 supervised training.}
    \label{tab:ablation_results}

\end{table}

 We conduct ablation studies to assess the effectiveness of pre-training DINOv2 on hand and face crops, the benefits of the multi-stream approach, and the potential of incorporating raw frames as an additional input stream.

\vspace{.1in}
\noindent\textbf{Do we benefit from pre-training face/hand specific DINOv2 feature extractors?}
As shown in Table~\ref{tab:ablation_results}, using the original DINOv2 features without further pre-training on hand and face crops achieves a BLEU score of 11.3. When we pre-train DINOv2 on hand and face crops and utilize the learned features, the BLEU score improves to 14.3, a substantial increase of 3.0 points. This indicates that while DINOv2 features are indeed robust, continued pre-training on domain-specific data (i.e., hand and face crops) enhances the model's ability to capture relevant information for sign language translation.

\noindent\textbf{Is the multi-stream approach beneficial compared to just using the original frames?}
To evaluate the effectiveness of the multi-stream approach, we compare the performance of using only the original frames (uncropped) with that of the multi-stream model. As shown in Table~\ref{tab:ablation_results}, the model trained on uncropped frames achieves a BLEU score of 4.9, while the multi-stream model (using the same original DINOv2 checkpoint model) obtains a significantly higher BLEU score of 11.4. This substantial improvement demonstrates the benefit of the multi-stream approach (given the same model as feature extractor) in capturing fine-grained details and relevant information from different body parts for sign language translation.

\noindent\textbf{Does a $5^{\textrm{th}}$ stream containing the raw frames help the model?}
We investigate the potential of incorporating an additional stream containing the raw frames alongside the upper body pose features and cropped hand and face features. As shown in Table~\ref{tab:ablation_results}, adding the global frame as an extra stream to the multi-stream model results in a BLEU score of 12.2, which is lower than the 14.3 BLEU score achieved by the model without the global frame. This suggests that the global frame may add noise rather than improve the model.

\section{Additional Analysis and Discussion}

\subsection{Scaling Analysis}
To better understand the scaling properties of our approach, we conduct additional experiments varying the amount of pre-training data. Table~\ref{tab:scaling_analysis} shows how model performance changes as we increase the number of frames used during pre-training from 0.1M to 1.0M frames on both How2Sign and YouTube-ASL datasets.

\begin{table}[t]
\centering
\begin{tabular}{l r r}
\toprule
Pre-training Frames & H2S & YT $\rightarrow$ H2S \\
\midrule
0.1M & 1.5 & 11.6 \\
0.4M & 1.5 & - \\
0.8M & 2.1 & - \\
1.0M & 2.4 & 14.3 \\
\bottomrule
\end{tabular}
\caption{Analysis of model performance (BLEU scores) with varying amounts of pre-training data.}
\label{tab:scaling_analysis}
\end{table}

The results demonstrate consistent improvement in translation quality as we increase the amount of pre-training data, with gains of 0.9 and 2.7 BLEU points on How2Sign and YouTube-ASL respectively when scaling from 0.1M to 1.0M frames. This suggests that our method can effectively leverage additional pre-training data, though the gains may be bounded by the model capacity of the ViT-Small architecture used in our experiments.

\subsection{Multi-Stream Contribution Analysis}
To validate the importance of each information stream in our multi-stream architecture, we conduct ablation experiments using different combinations of streams. Table~\ref{tab:stream_analysis} presents the results on the How2Sign dataset.

\begin{table}[t]
\centering
\begin{tabular}{l r}
\toprule
Streams Used & BLEU \\
\midrule
Face only & 0.6 \\
Hands only & 0.2 \\
Upper body only & 0.8 \\
Hands + Upper Body & 2.1 \\
All streams (full model) & 2.4 \\
\bottomrule
\end{tabular}
\caption{Impact of different information streams on translation performance (BLEU scores) on How2Sign.}
\label{tab:stream_analysis}
\end{table}

These results empirically demonstrate that each stream contributes meaningful information to the translation task. While the upper body stream provides the strongest individual signal (0.8 BLEU), the combination of streams yields substantially better performance (2.4 BLEU), supporting our hypothesis that different visual cues complement each other. The relatively low performance of individual streams also highlights the importance of our multi-stream approach in capturing the full range of linguistic information present in sign language videos.

\subsection{Choice of Visual Backbone}
We investigate different visual backbones for feature extraction. While prior work has established DINOv2's superiority over ResNet architectures for sign language tasks, we conducted additional experiments comparing DINOv2 with MAE on the How2Sign dataset.

\begin{table}[t]
\centering
\begin{tabular}{l r}
\toprule
Backbone & BLEU \\
\midrule
MAE (whole frame pre-trained) & 0.3 \\
DINOv2 (whole frame pre-trained) & 1.0 \\
DINOv2 (original) & 1.1 \\
\bottomrule
\end{tabular}
\caption{Comparison of visual backbones on How2Sign using a whole-frame approach.}
\label{tab:backbone_comparison}
\end{table}

The results, shown in Table~\ref{tab:backbone_comparison}, demonstrate DINOv2's effectiveness as a feature extractor for sign language translation. Notably, the performance difference between pre-trained and original DINOv2 is minimal when using whole frames, which motivated our exploration of the multi-stream approach.

\subsection{Data Efficiency and Scaling}
To understand the relationship between pre-training data volume and model performance, we evaluate our model at different stages of pre-training, as shown in Table~\ref{tab:data_scaling}.

\begin{table}[t]
\centering
\begin{tabular}{r r r}
\toprule
Pre-training data & Frames & BLEU \\
\midrule
10\% & 0.1M & 1.5 \\
100\% & 1.0M & 2.4 \\
\bottomrule
\end{tabular}
\caption{Impact of pre-training data volume on translation performance on How2Sign.}
\label{tab:data_scaling}
\end{table}

The improvement from 1.5 to 2.4 BLEU with increased pre-training data suggests that our approach can effectively utilize additional data. However, we note that these gains are achieved while using only 1.2\% of the available YouTube-ASL frames for pre-training, highlighting the efficiency of our method.

\section{Hyperparameter Settings}
\label{sec:hyperparameter}
The hyperparameter values used in our experiments are shown in~\Cref{tab:pretraining_settings} and~\Cref{tab:finetuning_settings_h2s_dm}.

\begin{table}[ht!]
\centering
\fontsize{10}{12}\selectfont
\sisetup{table-format = 3.2}
\resizebox{0.4\textwidth}{!}{%
\begin{tabular}{@{}ll@{}}
\toprule
\textsc{parameter}      & \textsc{value} \\ \midrule
dino: & \\
 loss\_weight & 1.0\\
 head\_nlayers & 3\\
 head\_hidden\_dim & 2048\\
 koleo\_loss\_weight & 0.1\\
ibot: & \\
 loss\_weight & 1.0\\
 mask\_sample\_probability & 0.5\\
 head\_bottleneck\_dim & 256\\
 head\_nlayers & 3\\
 head\_hidden\_dim & 2048\\
train: & \\
 batch\_size\_per\_gpu & 128\\
 saveckp\_freq & 20\\
 seed & 0\\
 num\_workers & 10\\
 OFFICIAL\_EPOCH\_LENGTH & 100\\
 cache\_dataset & true\\
 centering & sinkhorn\_knopp\\
student: & \\
 arch & vit\_small\\
 patch\_size & 14\\
 drop\_path\_rate & 0.4\\
 layerscale & 1.0e-05\\
 drop\_path\_uniform & false\\
 pretrained\_weights & null\\
 ffn\_layer & mlp\\
 block\_chunks & 0\\
 qkv\_bias & true\\
 proj\_bias & true\\
 ffn\_bias & true\\
 num\_register\_tokens & 4\\
 interpolate\_antialias & false\\
 interpolate\_offset & 0.1\\
teacher: & \\
 momentum\_teacher & 0.994\\
 final\_momentum\_teacher & 1\\
 warmup\_teacher\_temp & 0.04\\
 teacher\_temp & 0.07\\
 warmup\_teacher\_temp\_epochs & 30\\
 epochs & 100\\
 weight\_decay & 0.04\\
 weight\_decay\_end & 0.2\\
 base\_lr & 0.0002\\
 lr & 0.0\\
 warmup\_epochs & 10\\
 min\_lr & 1.0e-06\\
 clip\_grad & 3.0\\
 freeze\_last\_layer\_epochs & 1\\
 scaling\_rule & sqrt\_wrt\_1024\\
 patch\_embed\_lr\_mult & 0.2\\
 layerwise\_decay & 0.9\\
 adamw\_beta1 & 0.9\\
 adamw\_beta2 & 0.999\\
 crops: & \\
 global\_crops\_scale & (0.5,1.0)\\
 local\_crops\_number & 8\\
 local\_crops\_scale & (0.25,0.5)\\
 global\_crops\_size & 224\\
 local\_crops\_size & 98\\

\bottomrule
\end{tabular}%
}
\caption{SignMusketeers pre-training settings.}
\label{tab:pretraining_settings}
\end{table}

\begin{table*}[ht!]
\centering
\fontsize{10}{12}\selectfont
\sisetup{table-format = 3.2}
\resizebox{0.4\textwidth}{!}{%
\begin{tabular}{@{}ll@{}}
\toprule
\textsc{parameter}      & \textsc{value} \\ \midrule
 model\_name & t5-1.1\\
 target\_language & English\\
 
 expand\_time & false\\
 freeze\_model & false\\
 freeze\_adapter & false\\
 force\_target\_language & false\\
 batch\_size\_per\_gpu & 16\\
 learning\_rate & 0.001\\
 train\_steps & 5000\\
 fp16 & false\\
 test\_every & 500\\
 save\_every & 500\\
 log\_every & 500\\
 num\_workers & 4\\
 optimizer & adamw\_torch\\
 schedule & cosine\\
 weight\_decay & 1e-1\\
 label\_smoothing\_factor & 0.2\\
 generation\_max\_length & 128\\
 generation\_num\_beams & 5\\
 grad\_clipping & 1.0\\
\bottomrule
\end{tabular}%
}
\caption{Fine-tuning settings for How2Sign (H2S).}
\label{tab:finetuning_settings_h2s_dm}
\end{table*}

\section{MediaPipe Pose Landmark Indices}
\label{sec:mediapipe_pose_indices}

\begin{figure*}
    \centering
    \includegraphics[width=\textwidth]{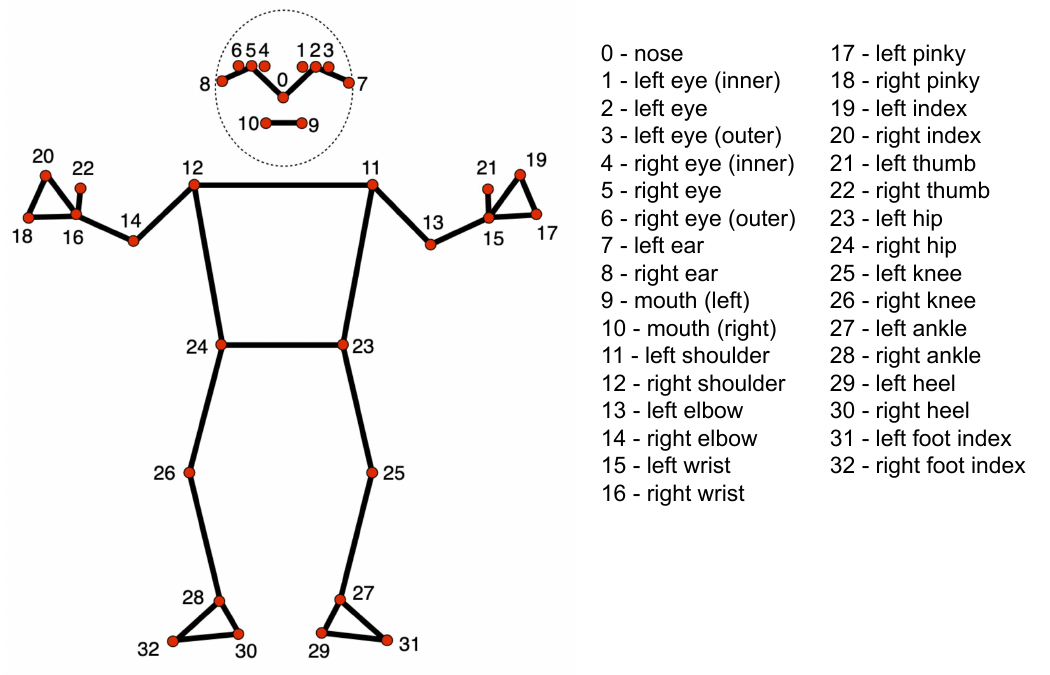}
    \caption{MediaPipe pose landmark indices. Source: \href{https://ai.google.dev/edge/mediapipe/solutions/vision/pose_landmarker\#pose_landmarker_model}{Google AI MediaPipe Documentation.}}
    \label{fig:mediapipe_indices}
\end{figure*}

The MediaPipe pose landmark indices are defined in~\Cref{fig:mediapipe_indices}.

\end{document}